\documentclass{article} % For LaTeX2e
\usepackage{iclr2027_conference,times}

\usepackage{amsmath,amsfonts,bm}

\def\eqref#1{equation~\ref{#1}}
\def\1{\bm{1}}

\DeclareMathAlphabet{\mathsfit}{\encodingdefault}{\sfdefault}{m}{sl}
\SetMathAlphabet{\mathsfit}{bold}{\encodingdefault}{\sfdefault}{bx}{n}

\usepackage{hyperref}
\usepackage{url}
\usepackage[table]{xcolor}
\usepackage{booktabs}

\usepackage{amsmath,amssymb}   % \substack, \text, \operatorname, \mathbb
\usepackage{graphicx}          % \resizebox
\usepackage{multirow}          % \multirow in tab:main_results
\usepackage{enumitem}          % [leftmargin=*]
\usepackage{algorithm}
\usepackage{algpseudocode}
\usepackage{subcaption} 

\definecolor{posgreen}{HTML}{008000}
\definecolor{negred}{HTML}{C00000}
\newcommand{\subpos}[1]{$_{\textcolor{posgreen}{\scriptstyle \boldsymbol{+#1}}}$}
\newcommand{\subneg}[1]{$_{\textcolor{negred}{\scriptstyle #1}}$}
\newcommand{\subneu}{$_{\textcolor{gray}{\scriptstyle \pm0.00}}$}

\newcommand{\subposgray}[1]{$_{\textcolor{gray}{\scriptstyle +#1}}$}
\newcommand{\subneggray}[1]{$_{\textcolor{gray}{\scriptstyle #1}}$}
\newcommand{\groupsep}{\arrayrulecolor{black!25}\cmidrule(l){2-9}\arrayrulecolor{black}}
\newcommand{\std}[1]{{\scriptsize$\,\pm\,$#1}}

\usepackage[T1]{fontenc}
\usepackage[most]{tcolorbox}   % 'most' also loads listings

\lstdefinestyle{prompt}{
  basicstyle=\ttfamily\footnotesize,
  breaklines=true, breakatwhitespace=true,
  columns=flexible, keepspaces=true,
  breakindent=0pt, postbreak={},
  showstringspaces=false,
  moredelim=[s][\color{teal!70!black}\bfseries]{\{}{\}},
}

\newtcblisting{promptbox}[2][gray]{
  enhanced, breakable, listing only,
  listing options={style=prompt},
  colback=#1!4, colframe=#1!55!black,
  boxrule=0.5pt, arc=2pt,
  left=4pt, right=4pt, top=2pt, bottom=2pt,
  fonttitle=\bfseries\small, title={#2},
}

\newcommand{\aff}[1]{\textsuperscript{\normalfont #1}}

\title{On-Policy Self-Training: Data-Free \\Self-Improvement in Large Language Models}

\iclrfinalcopy

\author{%
  \textbf{Leon Sick}\aff{1,2,$\ast$} \quad
  \textbf{Lukas Hoyer}\aff{1} \quad
  \textbf{Yannick Strümpler}\aff{1} \quad
  \textbf{Christian Reisswig}\aff{1} \\
  \textbf{Timo Ropinski}\aff{2} \quad
  \textbf{Federico Tombari}\aff{1} \\
  \aff{1}Google \qquad \aff{2}Ulm University
}

\begin{document}

\maketitle

{\renewcommand{\thefootnote}{$\ast$}\footnotetext{Work done during an internship at Google.}}

\begin{abstract}
On-policy self-distillation enables post-training foundation models without relying on stronger external teachers. However, existing approaches rely on ground-truth question-answer pairs, human-curated problem sets, or other supervised signals. In this work, we reformulate on-policy self-distillation as a self-training framework, driven by self-generated question-answer pairs and privileged reasoning demonstrations. We push supervision to the limit: guided solely by coarse, high-level topic descriptions, the model synthesizes its own curriculum of queries and privileged reasoning traces without access to any human-curated prompts or answers. We find that self-distillation on fully synthetic privileged information enables a model to improve across multiple domains while retaining its capabilities on domains it is not self-trained on. We demonstrate the generality of this framework across text and multimodal domains, achieving consistent self-improvement on standard language benchmarks and Visual Question Answering. Our findings demonstrate that foundation models possess rich latent capabilities that can be amplified without external task supervision or curated problem distributions.
\end{abstract}

\section{Introduction}

% - There are several approaches to post-train an LLM
% - SFT, RL, OPD/GKD, OPSD
% - They all require labeled data
% - Human labeled data is expensive to obtain
% - new line of work removing answer as supervision during post-training
% - however using GT answers requires human data curation & sets priors about which answers/learning signal to generate
% - we propose to go one step further, remove GT questions and generate data by only seeding the general topic in the generation prompt
% - Inspired by the sucessfull application of self-training in unsupervised computer vision, a setting where models generate their own data/annotations, we explore this same direction for LLMs
% - In our self-training loop, the model first receives a data generation prompt to generate Q/a pairs
% - we then experiment with various pseudo-verification approaches that serve as a quality proxy to filter this data
% - We explore the use of OPSD as the algorithm of choice to update the models parameters
% - We show that repeating this loop serveral times enables the LLM to continually self-improve

Post-training is a primary driver of capability gains in large language models \citep{ouyang2022training}, and its paradigms differ along two axes: whether training trajectories come from the model itself, and how dense the learning signal is. Supervised finetuning (SFT) imitates curated demonstrations, providing dense token-level targets on teacher-forced trajectories the model never visits at inference, which induces exposure bias \citep{gu2024minillm, agarwal2024policy} and erodes existing capabilities \citep{shenfeld2026rl}. Reinforcement learning with verifiable rewards (RLVR) trains on the model's own rollouts \citep{guo2025deepseek, shao2024deepseekmath}, avoiding this mismatch but compressing supervision into a single scalar per trajectory. On-policy distillation (OPD) combines the two: the student samples its own trajectories, and a teacher supplies dense next-token targets at the states it visits \citep{agarwal2024policy, gu2024minillm}. On-policy self-distillation (OPSD) removes the need for a stronger teacher by constructing teacher and student from the same parameters under asymmetric conditioning \citep{zhao2026self, shenfeld2026self}.
Across this progression, external supervision is relocated rather than removed: SFT requires reference solutions, RLVR gold answers, OPD a stronger teacher, and OPSD a privileged context, typically the ground-truth solution.

Curating this required supervision at scale is costly: process-level annotation of a single mathematics corpus required about 800,000 human step labels \citep{lightman2024let}, and open instruction datasets rest on crowdsourcing efforts spanning over 13,000 volunteers \citep{kopf2023openassistant}. This cost has motivated a shift toward synthetic supervision, in which the model generates its own instructions \citep{wang2023self}, rationales \citep{zelikman2022star}, or pseudo-solutions \citep{li2026uopsd}. However, these methods still start from human-written problems, so human curation still decides which problems the model trains on and, in turn, what it can learn.

In language models, self-training is an established approach to self-improvement \citep{zelikman2022star, huang2023large, singh2024beyond, zuo2026ttrl}. The model solves problems, filters its solutions by agreement with a reference answer or its own majority vote, and is finetuned on those it retains. These methods, however, still train on a curated problem set, which must be collected anew for every task \citep{singh2024beyond}. Yet a language model can pose problems, not only solve the ones it is given \citep{zhao2026absolute, chen2026sqlm}. Therefore, we investigate whether a language model can self-train without labels or a curated problem set, requiring only topic descriptors from which it generates its own training problems.

% In this work, we propose On-Policy Self-Training (OPST), which closes the self-training loop with dense supervision, generated by the model itself in the form of privileged information. Rather than assuming a distribution of human-written questions, we condition generation on a coarse topic descriptor alone, from which the model synthesizes both the query and the privileged information used to supervise it.

\begin{figure*}[t]
  \centering
  % (a) Self-training loop: page 1 of the PDF (narrow)
  \begin{minipage}[t]{0.60\textwidth}
    \centering
    \includegraphics[width=\linewidth,page=1]{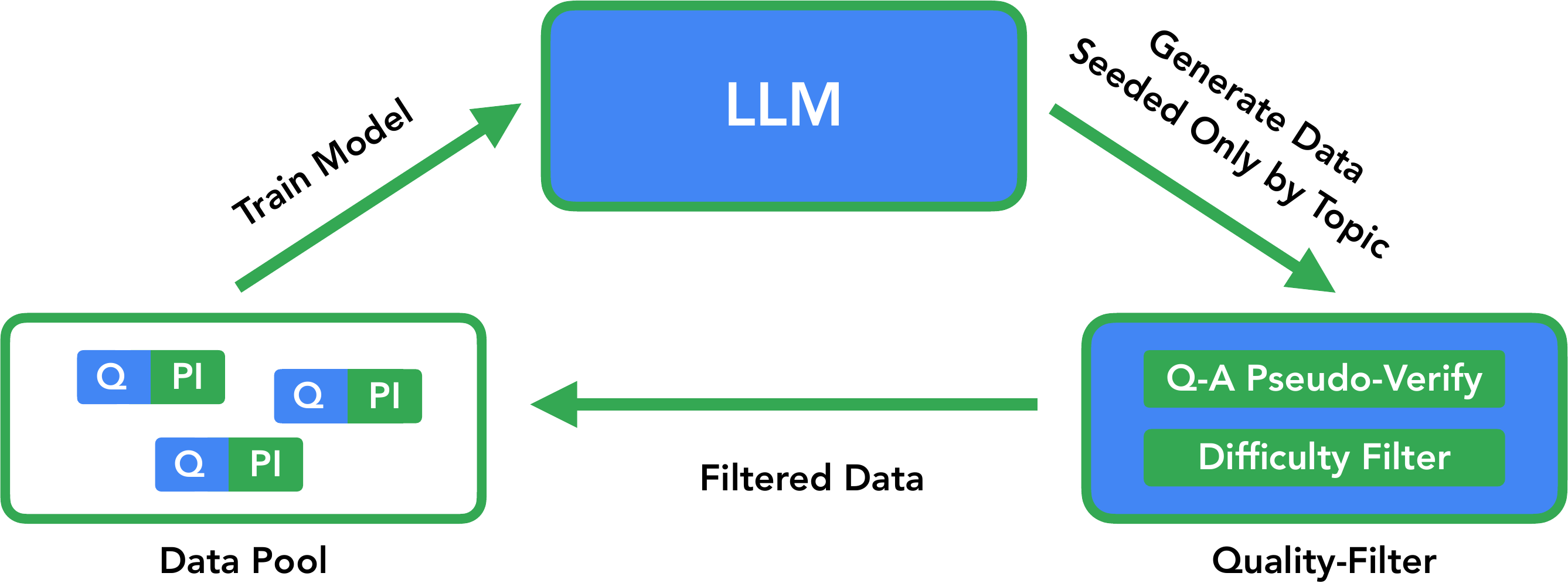}
    %\subcaption{On-Self-training loop}
    \label{fig:overview-loop-sub}
  \end{minipage}%\hfill
  % (b) Model training: page 2 of the PDF (wide)
  % \begin{minipage}[t]{0.40\textwidth}
  %   \centering
  %   \includegraphics[width=\linewidth,page=2]{opst_graphics_converted.pdf}
  %   \subcaption{Model training}
  %   \label{fig:overview-training}
  % \end{minipage}
  \vspace{-5mm}
  \caption{\textbf{On-Policy Self-Training loop.} Seeded only by topic descriptors, the model generates and filters its own question-answer pairs, then trains on them via on-policy self-distillation from a privileged EMA teacher, iterating for data-free self-improvement.}
  \vspace{-5mm}
  \label{fig:overview-loop}
\end{figure*}

\begin{figure*}[t]
  \centering
  % (a) Self-training loop: page 1 of the PDF (narrow)
  \begin{minipage}[t]{\textwidth}
    \centering
    \includegraphics[width=\linewidth,page=2]{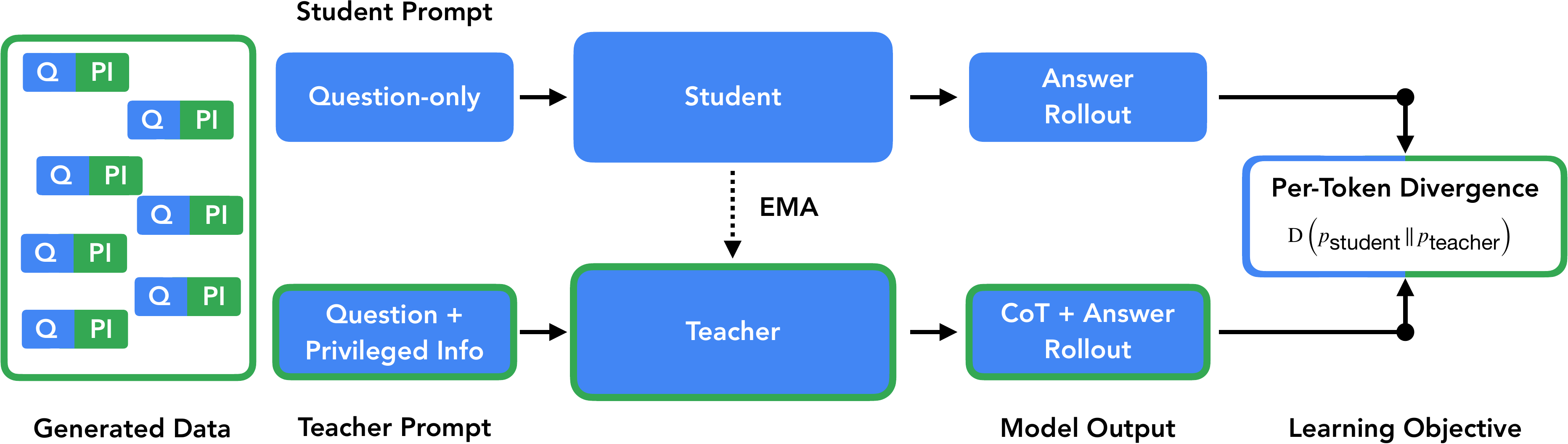}
    %\subcaption{On-Self-training loop}
    \label{fig:overview-training-sub}
  \end{minipage}%\hfill
  % (b) Model training: page 2 of the PDF (wide)
  % \begin{minipage}[t]{0.40\textwidth}
  %   \centering
  %   \includegraphics[width=\linewidth,page=2]{opst_graphics_converted.pdf}
  %   \subcaption{Model training}
  %   \label{fig:overview-training}
  % \end{minipage}
  \vspace{-5mm}
  \caption{\textbf{Model training.} As in OPSD \citep{zhao2026self}, the student sees only the question and matches the next-token distribution of a teacher that also sees privileged information. In OPST, this information is self-generated, and the teacher is an EMA of the student.}
  \vspace{-5mm}
  \label{fig:overview-training}
\end{figure*}

In this work, we propose \textbf{On-Policy Self-Training (OPST)}, an approach to self-train a language model with dense supervision derived from self-generated privileged information (Figure~\ref{fig:overview-loop}). We remove both ground-truth answers and the curated problem set: conditioned only on a coarse topic descriptor, the model synthesizes both a query and the privileged information (PI) used to train on it. Since the generated question-answer pairs are no longer human-curated, the quality of this data can be noisy. Hence, we filter the generated data along two axes: Pseudo-verification ensures that a problem is well-formed, requiring the model to reproduce its answer in a fresh context and under a paraphrase. Further, we apply a consensus gate to ensure that the problem is informative, discarding problems whose answer the student already consistently reproduces across on-policy rollouts. Because the gate is evaluated against the current student, the difficulty of the retained curriculum implicitly evolves as the student model improves. On the quality-filtered samples, we train the model with on-policy self-distillation (Figure~\ref{fig:overview-training}), where the teacher provides high-quality token-level supervision to the student, limiting catastrophic forgetting of existing capabilities acquired from supervised data \citep{shenfeld2026rl, shenfeld2026self}. These effects come from only the teacher being conditioned on the privileged information, while the student receives the question alone.

%This asymmetry is crucial in a purely synthetic setting. Alternative approaches, such as Supervised Finetuning (SFT) on the generated answer or Reinforcement Learning (RL) rewarding agreement with it, treat that answer as the optimization target, which bounds the supervision by the quality of the data generation: at best, the student learns to reproduce a terse and possibly incorrect answer. Instead, we show that if the generated answer is fed to the teacher as context instead, the teacher can expand it into the reasoning steps the generation omits, restructure it into a form the student can follow, and fall back on its parametric knowledge where the answer is incorrect. The student's supervision is therefore bounded not by the generated answer but by what the model can infer given that answer as priviledged information, which yields a richer and more error-tolerant signal than teacher forcing on the answer alone. Iterating this loop improves the model over successive rounds, while an exponential moving average (EMA) teacher stabilizes optimization against self-induced distribution drift. The information gap between teacher and student thus determines both the magnitude of the learning signal and which problems provide one. Our consensus gate exploits this property: on problems whose answer the student already reproduces consistently, conditioning the teacher on $z$ leaves its next-token distribution close to the student's, so the divergence these samples contribute to the objective is negligible.

This asymmetry is crucial in a purely synthetic setting. SFT on the generated answer, or RL rewarding agreement with it, treats the generated answer as the optimization target, which bounds the supervision by the quality of the generation. We show that providing the synthetic answer as context to the teacher instead enables it to produce a reasoning rollout from its own parametric knowledge that serves as an enriched learning signal to the student. The student's supervision is therefore extended by what the teacher can infer from the privileged, generated context, exceeding what the generated answer alone provides. %We further propose a consensus gate to retain only problems on which teacher and student remain sufficiently dissimilar, i.e. those whose answer the student does not yet reproduce consistently across its rollouts.
Iterating this loop improves the model over successive rounds, while an exponential moving average (EMA) teacher stabilizes against self-induced distribution drift.

%Removing all forms of human data curation raises the question of generated data quality. To mitigate this, we propose a suite of pseudo-verification mechanisms that act as a quality proxy to filter the resulting question-answer pairs. On the filtered samples, we adopt on-policy self-distillation to update the model, since training on self-generated data risks eroding the capabilities the model has previously acquired \citep{shenfeld2026rl, shenfeld2026self}. Here, teacher and student are initialized from the same model, but only the teacher is conditioned on the privileged information, while the student receives the question alone. Crucially, the teacher is not forced to reproduce the generated answer, but merely conditioned on it. It therefore remains free to enrich the supervision with intermediate reasoning steps the initially generated answer omits, and to fall back on its own knowledge where the generated privileged information is incorrect. Consequently, we extract a substantially richer and more error-tolerant learning signal than teacher forcing on the generated answer alone. Finally, we then iterate on this loop, allowing the model to continually improve without external task supervision.

In our experiments, we show that OPST improves language and multimodal reasoning without human-curated questions or answers, outperforming alternative training strategies. 
On Gemma~4 E2B~\citep{team2026gemma4}, self-training increases MMLU~\citep{hendrycks2021mmlu} accuracy by $+10.2$ points, with consistent gains on GSM8K~\citep{cobbe2021training}, GPQA~\citep{rein2023gpqa}, and HellaSwag~\citep{zellers2019hellaswag}. We also find that self-training in-domain transfers to out-of-distribution domains, with performance either remaining stable or also improving. Applied to vision-language models with unlabeled images alone, Visual OPST improves VQAv2 accuracy by $+6.9$ points.

\section{Related Work}

\noindent\textbf{On-policy self-distillation.} OPSD \citep{zhao2026self} constructs both policies from a single model, conditioning the teacher on a verified reference solution. Since on-policy methods train on the model's own samples, they forget less than SFT \citep{shenfeld2026rl}, and self-distillation has been shown to enable continual learning across multiple domains \citep{shenfeld2026self}. %In addition, because teacher and student share their weights, $q_t \approx p_t$ at positions where $z$ is uninformative. These positions contribute little to the loss, so the corresponding behavior remains largely unchanged.
Subsequent work varies the privileged context between expert demonstrations \citep{shenfeld2026self}, privileged observations \citep{penaloza2026privileged}, and environmental feedback \citep{huebotter2026rlsd}. Others address information leakage and instability \citep{yang2026self} or extend the setting to agentic tasks \citep{liu2026hero}. Closest to our work, \citet{li2026uopsd} replace the gold solution with a majority-vote consensus over the model's own rollouts, eliminating answer supervision. All of these, however, operate on a given distribution of problems. OPST removes that assumption: the model synthesizes its own queries from a coarse topic descriptor, so neither the problems nor their solutions are externally supplied.

\noindent\textbf{LLM Self-Training.} Early self-training methods filter solutions by agreement with a reference answer \citep{zelikman2022star, singh2024beyond} or by self-consistency across samples \citep{huang2023large}, and label-free reinforcement learning has since replaced the reference answer with majority votes \citep{zuo2026ttrl} or the model's own confidence \citep{zhao2026learning}. These methods remove answers but not problems: each still trains on a curated problem set, which must be collected anew for every task \citep{singh2024beyond}.
Self-play methods let a proposer generate tasks, seeded by as little as a topic description and checked by a code executor or by majority voting over the solver's answers \citep{zhao2026absolute, huang2026r, chen2026sqlm}. 
Self-generated problems, however, carry information that self-play reduces to a scalar reward: the answer and rationale the model produces while posing them. Used as privileged context for a self-distillation teacher, it instead yields dense, token-level supervision. On curated problems, \citet{li2026uopsd} show that self-generated answers are considerably more effective as teacher context than as a reward, and \citet{zhang2026piplay} combine self-play with self-distillation for search agents. While their method requires no external training data, question construction is grounded in documents retrieved from an external corpus, whereas OPST generates problems from the model's parametric knowledge alone. Whether a model can improve from problems it poses itself, with dense token-level supervision and without retrieval or other external grounding, has to our knowledge not yet been studied for general reasoning.

%They train the solver with reinforcement learning, and sparse self-generated rewards are prone to reward hacking and abrupt collapse \citep{shafayat2025selftrain}. 
%Without external grounding, proposer and solver also share the same knowledge, which limits what the solver can learn and lets errors in generated problems compound over iterations \citep{liu2025spice, shumailov2024collapse}.

\section{Method}
\label{sec:method}

In this section, we present On-Policy Self-Training. We first define the data-free post-training setting (Section~\ref{sec:problem_formulation}) and review on-policy self-distillation (Section~\ref{sec:preliminaries}). Following, we describe how the model generates problems and privileged information from topic descriptors (Section~\ref{sec:generation}), how it filters them intrinsically (Section~\ref{sec:verification}), and how it trains on the result (Section~\ref{sec:training}). Repeating these steps yields multiple rounds of self-training (Section~\ref{sec:self_training_loop}). Finally, we extend OPST to vision-language models (Section~\ref{sec:visual_opst}).

\subsection{Problem Formulation: Data-Free Post-Training}
\label{sec:problem_formulation}

Post-training typically relies on a curated dataset of human-written queries and ground-truth answers $\mathcal{D} = \{(x_i, y_i^*)\}$, or at least on a curated prompt distribution $\mathcal{D}_{\text{prompts}} = \{x_i\}$. In \emph{data-free post-training}, neither is available. Training is seeded only by a set of coarse topic descriptors $\mathcal{T} = \{\tau_1, \dots, \tau_K\}$, where each $\tau_k$ names a broad subject area, such as \emph{``Abstract Algebra''} or \emph{``Commonsense Causal Reasoning''}. From $\mathcal{T}$, the model generates a synthetic training set $\mathcal{D}_{\text{syn}} = \{(x_i, z_i)\}_{i=1}^N$, where $x_i$ is a generated query and $z_i$ its privileged information, consisting of a rationale and a candidate answer (Section~\ref{sec:generation}).

%In \emph{data-free post-training}, neither is available. Training is seeded only by a set of coarse topic descriptors
% \begin{equation}
%     \mathcal{T} = \{\tau_1, \tau_2, \dots, \tau_K\},
% \end{equation}
% where each $\tau_k$ names a broad subject area, such as \emph{``Abstract Algebra''} or \emph{``Commonsense Causal Reasoning''}. From $\mathcal{T}$, the model generates its own problems together with the supervision used to train on them, forming a synthetic training set
% \begin{equation}
%     \mathcal{D}_{\text{syn}} = \{(x_i, z_i)\}_{i=1}^N,
% \end{equation}
% where $x_i$ is a generated query and $z_i$ its privileged information, consisting of a rationale and a candidate answer (Section~\ref{sec:generation}).

The setting has two objectives. First, the model must improve its reasoning in the seeded domains using only self-generated problems and privileged information. Second, because the training signal comes from its own generations without human oversight, the model must retain its general reasoning and previously acquired parametric knowledge on domains it is not self-trained on, a prerequisite for repeating the process over multiple rounds.

\subsection{Preliminaries: On-Policy Self-Distillation}
\label{sec:preliminaries}

On-policy self-distillation~\citep{zhao2026self} derives the student $\pi_\theta$ and the teacher $\pi_\phi$ from the same model and replaces the capability gap of standard distillation \citep{kim2016sequence} with an information gap. Given a problem $x$, the student samples a response $y = (y_1, \dots, y_T) \sim \pi_\theta(\cdot \mid x)$, with next-token distribution
\begin{equation}
\label{eq:student_dist}
    p_t = \pi_\theta(\cdot \mid x, y_{<t})
\end{equation}
at step $t$. The teacher evaluates the same prefix $y_{<t}$, but is additionally conditioned on privileged information $z$ that the student does not see:
\begin{equation}
\label{eq:teacher_dist}
    q_t = \pi_\phi(\cdot \mid x, z, y_{<t}).
\end{equation}
The student is trained to match the teacher along its own trajectory,
\begin{equation}
\label{eq:opsd_loss}
    \mathcal{L}_{\text{OPSD}}(\theta) = \mathbb{E}_{(x, z) \sim \mathcal{D},\; y \sim \pi_\theta(\cdot \mid x)} \left[ \frac{1}{\sum_{t} M_t} \sum_{t=1}^{T} M_t \, D(p_t \,\|\, q_t) \right],
\end{equation}
where $D$ is a divergence between distributions, $M_t \in \{0, 1\}$ masks padding tokens, and no gradient flows through the teacher. In standard OPSD, $z$ is a reference solution \citep{zhao2026self}.

Existing OPSD methods, however, assume a curated problem set and typically use a reference solution as $z$. The following sections describe how OPST generates both with the model itself.

\subsection{Topic-Seeded Curriculum Generation}
\label{sec:generation}

Instead of drawing problems from a curated distribution, OPST generates them from the topic descriptors $\mathcal{T}$. Given a topic $\tau \in \mathcal{T}$, a generator $\pi_{\text{gen}}$ produces a query $x$ together with privileged information $z$,
\begin{equation}
\label{eq:synthesis}
    (x, z) \sim \pi_{\text{gen}}(\cdot \mid \tau), \quad z = (r, a),
\end{equation}
where $r$ is a rationale and $a$ a candidate answer. Queries can be multiple-choice or open-ended, and we sample at temperature $T_{\text{gen}}$ to obtain diverse problems. While the student receives the query $x$ alone, the teacher also receives $z$ as additional privileged information.  We embed the topic name into a targeted data generation prompt, asking the model to pose problems on the given topic along with generated privileged information. In practice, the model generates partial or full answers to the posed problem. Where available, topics are data-domain category labels.  The applied prompt templates are provided in Appendix~\ref{app:prompts}.

We instantiate $\pi_{\text{gen}}$ either as the EMA teacher $\pi_\phi$ (Section~\ref{sec:training}), so that data quality can improve as training progresses, or as a frozen copy of the initial model $\pi_{\theta_0}$. The student does not generate data, but its rollouts determine which samples are retained (Section~\ref{sec:verification}).

\subsection{Pseudo-Verification \& Consensus Gate}
\label{sec:verification}

% - Most informative samples are the ones where the student, across multiple rollouts, cannot arrive at a consistent answer
% - If the question is trivial to the student, it will produce the same answer every time
% - We propose to leverage this insight as a data filter which we call consensus gate
% - specifically, given a generated Q/A pair, we ask the the student to generate N rollouts. We measure the number of rollouts consistent with the generated answer. If all rollouts produce the same final answer, we discard the sample

Removing human supervision removes not only human data curation but also quality assurance. Generated samples can contain ill-posed questions or incorrect candidate answers, and training on them propagates these errors. At the same time, not every well-posed sample is worth training on: if the student already solves a problem reliably, the teacher conditioned on $z$ produces nearly the same next-token distribution as the student, and the sample contributes little beyond its generation and rollout cost.

We therefore filter any generated sample for two criteria: \emph{Pseudo-verification} first estimates whether a sample is well-formed, and the \emph{consensus gate} then approximates whether it is informative for the current student.

\paragraph{Pseudo-Verification.}
We evaluate if a given generated sample possesses two properties of a well-posed problem: its answer should be reproducible without the generator's own intent as context, and it should not depend on the particular wording of the question. %Throughout, $\hat{a}(q)$ denotes the answer the generator gives when solving a query $q$ at a low temperature $T_{\text{low}}$ without access to $z$.

To pseudo-verify the generated answer, we require the generator to recover it when solving the question in a fresh context, without access to $z$ and at a low temperature, which yields $\hat{a}(x) = a$. This separates authoring from solving: when query and answer are generated in one pass, a flawed or ambiguous question can still receive a confident answer.
To pseudo-verify the generated problem statement, we prompt the generator to paraphrase $x$ into $x' \sim \pi_{\text{gen}}(\cdot \mid x)$ with an instruction to preserve its meaning, solve $x'$ under the same conditions, and require the answer to be unchanged, $\hat{a}(x') = a$. A well-posed question has the same answer regardless of its wording, so this rejects questions whose answer depends on incidental phrasing, which typically indicates that the question admits more than one reading. A sample is retained only if: $v(x, z) = \mathbb{I}\big[\hat{a}(x) = a\big] \cdot \mathbb{I}\big[\hat{a}(x') = a\big]$

%The two conditions probe different failure modes but share a limitation: both measure self-consistency rather than correctness, so an error the model makes consistently passes them. Pseudo-verification reduces noise in the generated data without removing it, which is why we train with an objective that tolerates the inaccuracies that remain (Section~\ref{sec:training}).

\paragraph{Consensus Gate.}
We assume that the samples that carry the most signal are those the student cannot yet answer reliably: a question that is trivial for the student yields the same answer on every rollout, and the privileged teacher has little to add. We therefore measure answer agreement across sampled reasoning paths as in self-consistency \citep{wang2023selfconsistency}. For each verified sample, the student samples $G$ rollouts $\{y_g\}_{g=1}^{G} \sim \pi_\theta(\cdot \mid x)$, from which we extract answers $\{\hat{a}_g\}_{g=1}^{G}$. Samples with full agreement with the candidate answer, $c(x, z) = \frac{1}{G} \sum_{g=1}^{G} \mathbb{I}[\hat{a}_g = a] = 1$, provide little new signal for the model and are discarded.

%The two filters act in opposite directions by design. Pseudo-verification keeps samples with answers the model can reproduce under a fresh context and a paraphrased problem, while our consensus gate discards samples the student can already confidently answer. Samples that are retained target at the student's capability frontier: the problem can be considered stable enough for the generated answer to serve as usable privileged information, but the student does not yet solve it reliably, so the information gap between teacher and student is largest for exactly the retained samples. Since the consensus gate is a direct result of the student's capabilities, our filter evolves with the self-training loop. Synthetic samples that become trivial as the student improves are no longer part of the training set, so the difficulty of the retained curriculum implicitly evolves with the model rather than following a fixed distribution.
The two filters act in opposite directions by design. Pseudo-verification keeps samples that are stable enough for the generated answer to serve as privileged information, while the consensus gate discards samples the student already solves. The retained samples therefore lie at the student's capability frontier, and since the gate depends on the current student, the difficulty of the curriculum evolves with the model.

\subsection{Model Training}
\label{sec:training}

Given a pseudo-verified batch $\mathcal{B} = \{(x_i, z_i)\}_{i=1}^B$, we update the student with on-policy self-distillation (Section~\ref{sec:preliminaries}), where $p_{i,t}$ and $q_{i,t}$ denote the student and teacher distributions (Eqs.~\ref{eq:student_dist} and~\ref{eq:teacher_dist}) at step $t$ of the student's rollout on sample $i$.

\paragraph{Generalized Divergence.}
Following the generalized formulation of prior work \citep{agarwal2024policy, zhao2026self}, we instantiate the divergence $D$ in Eq.~(\ref{eq:opsd_loss}) as a skewed Jensen--Shannon divergence. With the interpolated mixture distribution $\bar{q}_{i,t} = 1-\alpha)\, q_{i,t} + \alpha\, p_{i,t}$ for $\alpha \in (0,1)$, the per-token divergence is defined as
\begin{equation}
    D_\alpha(p_{i,t} \,\|\, q_{i,t}) = \alpha\, D_{\mathrm{KL}}(q_{i,t} \,\|\, \bar{q}_{i,t}) + (1-\alpha)\, D_{\mathrm{KL}}(p_{i,t} \,\|\, \bar{q}_{i,t}).
\end{equation}
In the limits, $D_\alpha / \alpha \to D_{\mathrm{KL}}(q_{i,t} \,\|\, p_{i,t})$ as $\alpha \to 0$, recovering the mode-covering forward KL divergence, and $D_\alpha / (1-\alpha) \to D_{\mathrm{KL}}(p_{i,t} \,\|\, q_{i,t})$ as $\alpha \to 1$, recovering the mode-seeking reverse KL divergence. Unlike unbounded KL objectives, $D_\alpha$ is upper-bounded by the binary entropy $H(\alpha)$, ensuring that tokens to which one distribution assigns near-zero probability, such as localized artifacts introduced by an imperfect synthetic hint, are prevented from dominating the gradient update. We set $\alpha = 0.8$, which favors the mode-seeking regime. The final objective is
\begin{equation}
\label{eq:opst_loss}
    \mathcal{L}_{\text{OPST}}(\theta; \mathcal{B}) = \frac{1}{B} \sum_{i=1}^B \, \frac{1}{\sum_{t} M_{i,t}} \sum_{t=1}^{T} M_{i,t} \, D_\alpha(p_{i,t} \,\|\, q_{i,t}).
\end{equation}

\paragraph{EMA Teacher.}
The teacher is initialized from the student, $\phi_0 = \theta_0$, and updated after every gradient step as $\phi_k = (1-\beta)\,\phi_{k-1} + \beta\,\theta_k$. Since the teacher produces the dense supervision, it must remain stable. With $\beta = 1$, a single noisy batch immediately changes the targets for subsequent batches, and errors can reinforce themselves. The EMA instead averages out such fluctuations over roughly $1/\beta$ steps. Averaging the student's weights also yields a more accurate model than the student itself (Table~\ref{tab:component_ablations}d), consistent with momentum teachers in vision \citep{caron2021dino}. For this reason, the teacher is the model we evaluate. %For the same reason, the EMA teacher is the model we evaluate. When it also serves as the generator (Section~\ref{sec:generation}), the data of later rounds is generated and verified by the smoothed model rather than the noisy student.

\subsection{Iterative Self-Training}
\label{sec:self_training_loop}

A single round of OPST is limited by the quality of the data and supervision that the initial model provides. We therefore repeat generation, verification, and training over $N$ rounds. The student and teacher weights carry over from one round to the next. For each additional round,  $\pi_{\text{gen}}$ generates a new training set, from which we remove problems that duplicate those of earlier rounds.

The iterative nature of our loop lets student and teacher co-evolve. Training on new data further improves the student, and through the EMA update, also the teacher. A stronger teacher, in turn, benefits the next round in three ways. It provides more accurate token-level targets on the student's rollouts. As the generator, it produces better problems and privileged information. And as the verifier (Section~\ref{sec:verification}), it judges its own data more reliably, so fewer incorrect samples reach training. Because the teacher is an average over many training steps, only improvements that persist across batches enter this cycle. 

% \begin{algorithm}[t]
% \caption{On-Policy Self-Training (OPST)}
% \label{alg:opst}
% \begin{algorithmic}[1]
% \Require topics $\mathcal{T}$, initial weights $\theta_0$, rounds $N$, EMA rate $\beta$
% \State $\theta \gets \theta_0$, \ $\phi \gets \theta_0$
% \For{$n = 1, \dots, N$}
%     \State Generate $(x_i, z_i) \sim \pi_{\text{gen}}(\cdot \mid \tau)$ for $\tau \in \mathcal{T}$ \Comment{\S\ref{sec:generation}}
%     \State Pseudo-Verification; discard noisy samples and duplicates \Comment{Section~\ref{sec:verification}}
%     \For{each mini-batch $\mathcal{B}$}
%         \State Sample rollouts $y_i \sim \pi_\theta(\cdot \mid x_i)$ for all samples in $\mathcal{B}$
%         \State Update $\theta$ with $\nabla_\theta \mathcal{L}_{\text{OPST}}(\theta; \mathcal{B})$ \Comment{Eq.~\ref{eq:opst_loss}}
%         \State $\phi \gets (1-\beta)\,\phi + \beta\,\theta$ 
%     \EndFor
% \EndFor
% \State \Return $\pi_\phi$
% \end{algorithmic}
% \end{algorithm}

\subsection{Visual OPST}
\label{sec:visual_opst}

OPST is modality-agnostic and can be extended to the multimodal setting by adding an unlabeled modality. We therefore explore this for vision-language tasks by adding images to the process. Visual OPST uses a collection of unlabeled images $\mathcal{I}$ and visual topic descriptors $\mathcal{T}_{\text{vis}}$, such as \emph{``Spatial Relations''}, \emph{``Object Counting''}, or \emph{``Attribute Recognition''}. Unlike the text setting, it requires unlabeled images as input, from which we generate the same data as detailed in Section~\ref{sec:generation}.

\paragraph{Visual Curriculum Generation.}
Given an image $I \in \mathcal{I}$ and a topic $\tau \in \mathcal{T}_{\text{vis}}$, the generator produces a question about the image together with privileged information, $(x, z) \sim \pi_{\text{gen}}(\cdot \mid I, \tau)$. We consider two forms of privileged information. In the first,  $z$ is a detailed scene caption that the generator writes for $I$, describing its objects, attributes, and spatial arrangement. In the second, we prompt the model to detect the objects in the scene, from which we bootstrap bounding-box grounded spatial relations. The privileged caption informs the teacher rollout with explicit visual details that the student must extract from the image itself, while the detected spatial relations help ground the teacher with actual object positions. Both forms can be combined by conditioning the teacher on the caption and the spatial relations.

\paragraph{Visual Verification.}
Visual verification tests whether the candidate answer is supported by the image rather than by a hallucinated premise. The generator solves $x$ on $K$ augmented views $\mathcal{A}_k(I)$, obtained with color jitter, rescaling, and random cropping, and we measure agreement with the candidate answer:
\begin{equation}
    v_{\text{aug}}(I, x, z) = \frac{1}{K} \sum_{k=1}^{K} \mathbb{I}\big[\hat{a}(\mathcal{A}_k(I), x) = a\big].
\end{equation}
Horizontal flips are excluded because they change the answer to questions about left and right. %Unlike the binary checks of Section~\ref{sec:verification}, $v_{\text{aug}}$ yields a soft weight.
% TODO: add spatial self-grounding and question-answer cycle consistency only if they were used in the reported results.

\paragraph{Model Training.}
Training follows Eq.~(\ref{eq:opst_loss}) with the image added to both contexts: the student samples $y_i \sim \pi_\theta(\cdot \mid I_i, x_i)$, and the teacher provides $q_{i,t} = \pi_\phi(\cdot \mid I_i, x_i, z_i, y_{i,<t})$.

\section{Experiments}
\label{sec:experiments}

Our experimental evaluation focuses on three aspects: in-domain self-improvement without human-curated data, knowledge retention on out-of-domain (OOD) benchmarks when mixing data generation domains, and transfer to multimodal inputs with Visual OPST.

\subsection{Experimental Setup}
\label{sec:exp_setup}

\paragraph{Models and Datasets.}
As base models, we utilize the Gemma~4 E2B and E4B models~\citep{team2026gemma4}. %Prior self-play methods often start from pre-trained base models \citep{zhao2026absolute, huang2026r}, where learning the answer format alone yields large gains. we tarting from instruction-tuned checkpoints requires improvements beyond formatting. 
Data Generation is seeded with topic descriptors covering the domains of four \emph{in-domain} benchmarks: MMLU \citep{hendrycks2021mmlu}, GPQA \citep{rein2023gpqa}, GSM8K \citep{cobbe2021training}, and HellaSwag \citep{zellers2019hellaswag}. To measure transfer and retention, we additionally evaluate on three \emph{OOD} benchmarks that no topic descriptor targets: ARC-Challenge \citep{clark2018think}, MATH-500 \citep{lightman2024let, hendrycks2021math500}, and Winogrande \citep{sakaguchi2021winogrande}. For the multi-domain experiments (Table~\ref{tab:domain_mixing}), GPQA is treated as out-of-domain. %We train for $N = 20$ rounds with batch size $B = 2$, learning rate $5 \times 10^{-6}$, EMA rate $\beta = 0.05$, and generation temperature $T_{\text{gen}} = 1.0$. Topic descriptors and further details are provided in Appendix~\ref{app:implementation}..

\paragraph{Evaluation and Baselines.}
Table~\ref{tab:main_results} compares training algorithms on identical self-generated data. \textit{SFT} imitates the generated answer, \textit{GRPO} \citep{shao2024deepseekmath} rewards agreement with it, and OPST supplies it as teacher context. Table~\ref{tab:supervision_levels} varies the level of supervision. \textit{Supervised SFT} and \textit{Supervised OPSD} train on curated questions with reference solutions, \textit{u-OPSD} \citep{li2026uopsd} on curated questions with self-generated answers, and \textit{R-Zero} \citep{huang2026r} on self-generated problems with reinforcement learning. \textit{Base} denotes the instruction-tuned model. In Table~\ref{tab:supervision_levels}, methods above the light rule train on human-curated reference solutions, and \textbf{bold} marks the best result among the remaining methods. Elsewhere, \textbf{bold} and \underline{underline} mark the best and second-best result. For OPST, we evaluate the EMA teacher (Section~\ref{sec:training}).

% \subsection{In-Domain Results}
% \label{sec:main_results}

% \begin{table*}[t]
% \centering
% \small
% \caption{\textbf{In-domain reasoning performance.} We report Pass@1 and Pass@5 accuracy (\%) across all four reasoning benchmarks. Without access to human-curated questions or ground-truth answers, OPST substantially improves accuracy over the instruction-tuned base model.}
% \label{tab:main_results}
% \vspace{2mm}
% \resizebox{\textwidth}{!}{%
% \setlength{\tabcolsep}{5.5pt}
% \begin{tabular}{l cc cc cc cc cc}
% \toprule
% & \multicolumn{2}{c}{\textbf{MMLU}} & \multicolumn{2}{c}{\textbf{GPQA}} & \multicolumn{2}{c}{\textbf{GSM8K}} & \multicolumn{2}{c}{\textbf{HellaSwag}} & \multicolumn{2}{c}{\textbf{Average}} \\
% \cmidrule(lr){2-3} \cmidrule(lr){4-5} \cmidrule(lr){6-7} \cmidrule(lr){8-9} \cmidrule(lr){10-11}
% \textbf{Method} & \textbf{Pass@1} & \textbf{Pass@5} & \textbf{Pass@1} & \textbf{Pass@5} & \textbf{Pass@1} & \textbf{Pass@5} & \textbf{Pass@1} & \textbf{Pass@5} & \textbf{Pass@1} & \textbf{Pass@5} \\
% \midrule
% Base Model (Gemma~4 E2B IT) & 37.46 & -- & 28.79 & -- & 73.44 & -- & 48.73 & -- & 47.11 & -- \\
% \rowcolor{gray!12}
% \textbf{OPST (Ours)} & \textbf{47.66} & \textbf{--} & \textbf{32.83} & \textbf{--} & \textbf{78.91} & \textbf{--} & \textbf{50.68} & \textbf{--} & \textbf{52.52} & \textbf{--} \\
% \midrule
% \textit{Gain ($\Delta$)} & \textit{+10.20} & \textit{--} & \textit{+4.04} & \textit{--} & \textit{+5.47} & \textit{--} & \textit{+1.95} & \textit{--} & \textit{+5.41} & \textit{--} \\
% \bottomrule
% \end{tabular}%
% }
% \end{table*}

\begin{table*}[t]
\centering
\small
\caption{\textbf{Synthetic-data results.} Performance comparison on language reasoning benchmarks for Gemma 4 models trained on self-generated data only.}
\label{tab:main_results}
\vspace{-2mm}
\resizebox{\textwidth}{!}{%
\setlength{\tabcolsep}{6pt}
\begin{tabular}{l l cc ccccc}
\toprule
& \textbf{Method} & \textbf{Signal} & \textbf{Gen. Answer} & \textbf{MMLU} & \textbf{GPQA} & \textbf{GSM8K}$^\ast$ & \textbf{HellaSwag} & \textbf{Average} \\
\midrule
\multicolumn{9}{l}{\textit{Gemma 4 E2B}} \\
& Base & & & 37.5 & 28.8 & -- & 48.7 & 38.3 \\
& + SFT & Dense & Target & \underline{38.3}\subposgray{0.8} & \underline{30.8}\subposgray{2.0} & -- & \underline{50.8}\subposgray{2.1} & \underline{40.0}\subposgray{1.7} \\
& + GRPO & Sparse & Reward & 37.3\subneggray{-0.2} & 30.3\subposgray{1.5} & -- & 50.3\subposgray{1.6} & 39.3\subposgray{1.0} \\
\rowcolor{gray!12}
\multirow{-4}{*}{\textit{w/o Think.}} & \textbf{+ OPST (Ours)} & Dense & Context & \textbf{47.7}\subpos{10.2} & \textbf{32.8}\subpos{4.0} & -- & \textbf{56.2}\subpos{7.5} & \textbf{45.6}\subpos{7.3} \\
\cmidrule(l){2-9}
& Base & & & 69.9 & 27.8 & 72.7 & 63.1 & 58.4 \\
& + SFT & Dense & Target & \underline{71.0}\subposgray{1.1} & 30.3\subposgray{2.5} & \underline{75.4}\subposgray{2.7} & \underline{64.5}\subposgray{1.4} & \underline{60.3}\subposgray{1.9} \\
& + GRPO & Sparse & Reward & 70.7\subposgray{0.8} & \underline{32.3}\subposgray{4.5} & 75.0\subposgray{2.3} & 62.5\subneggray{-0.6} & 60.1\subposgray{1.7} \\
\rowcolor{gray!12}
\multirow{-4}{*}{\textit{w/ Think.}} & \textbf{+ OPST (Ours)} & Dense & Context & \textbf{71.1}\subpos{1.2} & \textbf{34.3}\subpos{6.5} & \textbf{80.9}\subpos{8.2} & \textbf{65.6}\subpos{2.5} & \textbf{63.0}\subpos{4.6} \\
\midrule
\multicolumn{9}{l}{\textit{Gemma 4 E4B}} \\
& Base & & & 54.3 & 30.3 & -- & 53.7 & 46.1 \\
& + SFT & Dense & Target & \underline{58.1}\subposgray{3.8} & \underline{35.9}\subposgray{5.6} & -- & \textbf{58.6}\subposgray{4.9} & \underline{50.9}\subposgray{4.8} \\
& + GRPO & Sparse & Reward & 55.8\subposgray{1.5} & 31.8\subposgray{1.5} & -- & 54.3\subposgray{0.6} & 47.3\subposgray{1.2} \\
\rowcolor{gray!12}
\multirow{-4}{*}{\textit{w/o Think.}} & \textbf{+ OPST (Ours)} & Dense & Context & \textbf{61.9}\subpos{7.6} & \textbf{36.4}\subpos{6.1} & -- & \underline{58.2}\subpos{4.5} & \textbf{52.2}\subpos{6.1} \\
\cmidrule(l){2-9}
& Base & & & 76.0 & 43.9 & 84.4 & 80.9 & 71.3 \\
& + SFT & Dense & Target & \underline{76.2}\subposgray{0.2} & 44.4\subposgray{0.5} & \underline{87.1}\subposgray{2.7} & 81.1\subposgray{0.2} & \underline{72.2}\subposgray{0.9} \\
& + GRPO & Sparse & Reward & 75.4\subneggray{-0.6} & \underline{45.0}\subposgray{1.1} & 83.2\subneggray{-1.2} & \underline{81.5}\subposgray{0.6} & 71.3\subneu \\
\rowcolor{gray!12}
\multirow{-4}{*}{\textit{w/ Think.}} & \textbf{+ OPST (Ours)} & Dense & Context & \textbf{80.1}\subpos{4.1} & \textbf{48.0}\subpos{4.1} & \textbf{93.0}\subpos{8.6} & \textbf{83.0}\subpos{2.1} & \textbf{76.0}\subpos{4.7} \\
\bottomrule
\multicolumn{9}{l}{\footnotesize $^\ast$Evaluated with thinking only} \\
\end{tabular}%
}
\end{table*}

\begin{table*}[t]
\centering
\small
\caption{\textbf{Results across various levels of supervision.} Performance comparison on language reasoning benchmarks for Gemma 4 models across supervision levels. Methods above the light rule train on human-curated reference solutions. \textbf{Bold} marks the best result among the remaining methods.}
\label{tab:supervision_levels}
\vspace{-2mm}
\resizebox{\textwidth}{!}{%
\setlength{\tabcolsep}{6pt}
\begin{tabular}{l l cc ccccc}
\toprule
& & \multicolumn{2}{c}{\textbf{Human-Curated}} & & & & & \\
\cmidrule(lr){3-4}
& \textbf{Method} & \textbf{Question} & \textbf{Answer} & \textbf{MMLU} & \textbf{GPQA} & \textbf{GSM8K}$^\ast$ & \textbf{HellaSwag} & \textbf{Average} \\
\midrule
\multicolumn{9}{l}{\textit{Gemma 4 E2B}} \\
& Base & & & 37.5 & 28.8 & -- & 48.7 & 38.3 \\
& + Supervised SFT & $\checkmark$ & $\checkmark$ & 47.3\subposgray{9.8} & 31.3\subposgray{2.5} & -- & 52.8\subposgray{4.1} & 43.8\subposgray{5.5} \\
& + Supervised OPSD & $\checkmark$ & $\checkmark$ & 49.6\subposgray{12.1} & 33.3\subposgray{4.5} & -- & 56.8\subposgray{8.1} & 46.6\subposgray{8.3} \\
\groupsep
& + u-OPSD$^\ddagger$ & $\checkmark$ & $\times$ & 45.9\subposgray{8.4} & \textbf{33.3}\subposgray{4.5} & -- & 54.7\subposgray{6.0} & 44.6\subposgray{6.3} \\
& + R-Zero$^\ddagger$ & $\times$ & $\times$ & 43.5\subposgray{6.0} & 30.8\subposgray{2.0} & -- & 51.2\subposgray{2.5} & 41.8\subposgray{3.5} \\
\rowcolor{gray!12}
\multirow{-6}{*}{\textit{w/o Think.}} & \textbf{+ OPST (Ours)} & $\times$ & $\times$ & \textbf{47.7}\subpos{10.2} & 32.8\subpos{4.0} & -- & \textbf{56.2}\subpos{7.5} & \textbf{45.6}\subpos{7.3} \\
\cmidrule(l){2-9}
& Base & & & 69.9 & 27.8 & 72.7 & 63.1 & 58.4 \\
& + Supervised SFT & $\checkmark$ & $\checkmark$ & 70.5\subposgray{0.6} & 30.8\subposgray{3.0} & 83.6\subposgray{10.9} & 63.7\subposgray{0.6} & 62.2\subposgray{3.8} \\
& + Supervised OPSD & $\checkmark$ & $\checkmark$ & 70.1\subposgray{0.2} & 31.8\subposgray{4.0} & 85.3\subposgray{12.6} & 63.9\subposgray{0.8} & 62.8\subposgray{4.4} \\
\groupsep
& + u-OPSD$^\ddagger$ & $\checkmark$ & $\times$ & 70.8\subposgray{0.9} & 33.3\subposgray{5.5} & 76.6\subposgray{3.9} & 65.0\subposgray{1.9} & 61.4\subposgray{3.0} \\
& + R-Zero$^\ddagger$ & $\times$ & $\times$ & 70.3\subposgray{0.4} & 31.8\subposgray{4.0} & 80.3\subposgray{7.6} & 64.1\subposgray{1.0} & 61.6\subposgray{3.2} \\
\rowcolor{gray!12}
\multirow{-6}{*}{\textit{w/ Think.}} & \textbf{+ OPST (Ours)} & $\times$ & $\times$ & \textbf{71.1}\subpos{1.2} & \textbf{34.3}\subpos{6.5} & \textbf{80.9}\subpos{8.2} & \textbf{65.6}\subpos{2.5} & \textbf{63.0}\subpos{4.6} \\
\midrule
\multicolumn{9}{l}{\textit{Gemma 4 E4B}} \\
& Base & & & 54.3 & 30.3 & -- & 53.7 & 46.1 \\
& + Supervised SFT & $\checkmark$ & $\checkmark$ & 62.4\subposgray{8.1} & 34.3\subposgray{4.0} & -- & 60.3\subposgray{6.6} & 52.3\subposgray{6.2} \\
& + Supervised OPSD & $\checkmark$ & $\checkmark$ & 63.4\subposgray{9.1} & 36.9\subposgray{6.6} & -- & 62.6\subposgray{8.9} & 54.3\subposgray{8.2} \\
\groupsep
& + u-OPSD$^\ddagger$ & $\checkmark$ & $\times$ & 60.0\subposgray{5.7} & 32.8\subposgray{2.5} & -- & 56.5\subposgray{2.8} & 49.8\subposgray{3.7} \\
& + R-Zero$^\ddagger$ & $\times$ & $\times$ & 54.7\subposgray{0.4} & 31.3\subposgray{1.0} & -- & 54.9\subposgray{1.2} & 47.0\subposgray{0.9} \\
\rowcolor{gray!12}
\multirow{-6}{*}{\textit{w/o Think.}} & \textbf{+ OPST (Ours)} & $\times$ & $\times$ & \textbf{61.9}\subpos{7.6} & \textbf{36.4}\subpos{6.1} & -- & \textbf{58.2}\subpos{4.5} & \textbf{52.2}\subpos{6.1} \\
\cmidrule(l){2-9}
& Base & & & 76.0 & 43.9 & 84.4 & 80.9 & 71.3 \\
& + Supervised SFT & $\checkmark$ & $\checkmark$ & 78.5\subposgray{2.5} & 43.5\subneggray{-0.4} & 93.2\subposgray{8.8} & 81.5\subposgray{0.6} & 74.2\subposgray{2.9} \\
& + Supervised OPSD & $\checkmark$ & $\checkmark$ & 77.7\subposgray{1.7} & 47.5\subposgray{3.6} & 93.4\subposgray{9.0} & 82.4\subposgray{1.5} & 75.3\subposgray{4.0} \\
\groupsep
& + u-OPSD$^\ddagger$ & $\checkmark$ & $\times$ & 76.9\subposgray{0.9} & 44.3\subposgray{0.4} & 88.9\subposgray{4.5} & 82.8\subposgray{1.9} & 73.2\subposgray{1.9} \\
& + R-Zero$^\ddagger$ & $\times$ & $\times$ & 75.4\subneggray{-0.6} & 44.0\subposgray{0.1} & 90.2\subposgray{5.8} & 82.1\subposgray{1.2} & 72.9\subposgray{1.6} \\
\rowcolor{gray!12}
\multirow{-6}{*}{\textit{w/ Think.}} & \textbf{+ OPST (Ours)} & $\times$ & $\times$ & \textbf{80.1}\subpos{4.1} & \textbf{48.0}\subpos{4.1} & \textbf{93.0}\subpos{8.6} & \textbf{83.0}\subpos{2.1} & \textbf{76.0}\subpos{4.7} \\
\bottomrule
\multicolumn{9}{l}{\footnotesize $^\ast$Evaluated with thinking only. $^\ddagger$Our re-implementation.} \\
\end{tabular}%
}
\end{table*}

\subsection{Results}
\label{sec:results}

\paragraph{Learning from Self-Generated Data.}
On Gemma~4 E2B without thinking, OPST improves MMLU by $+10.2$ points, GPQA by $+4.0$, and HellaSwag by $+7.5$ (Table~\ref{tab:main_results}). On average, OPST exceeds SFT on the same data by $5.6$/$1.3$ points (2B/4B) without thinking and by $2.7$/$3.8$ points with thinking. The two methods differ only in whether the generated answer is the optimization target or the teacher's context, so this gap isolates the effect of that choice. Only on 4B HellaSwag without thinking is SFT ahead ($0.4$ points). With thinking, the stronger base models leave less headroom on MMLU ($+1.2$ at 2B), whereas the gains on GPQA ($+6.5$) and GSM8K ($+8.2$) persist.

\textbf{Comparison Across Supervision Levels.}
Table~\ref{tab:supervision_levels} compares OPST against methods that use human-curated data. Among methods without reference solutions, OPST attains the best average in all four settings, exceeding u-OPSD by $1.0$ to $2.8$ points and R-Zero by $1.4$ to $5.2$ points. As u-OPSD trains on curated questions and R-Zero operates at the same level of supervision as OPST, this suggests that a curated problem set is not necessary and that the improvements stem from our training approach rather than additional data. OPST also closes most of the gap to full supervision: at 2B without thinking, it recovers $88\%$ of the supervised OPSD gain ($+7.3$ of $+8.3$), and with thinking, it matches supervised OPSD at both model sizes ($+0.2$ and $+0.7$ points). %Only at 4B without thinking do the supervised baselines remain ahead.

\begin{table*}[t]
\centering
\small
\caption{\textbf{Multi-domain results.} Performance comparison on in-domain and out-of-domain benchmarks for Gemma 4 E2B self-trained on combinations of topic descriptors. \textit{Single domain} reports separate models, each self-trained only on the topics of the respective benchmark.}
\label{tab:domain_mixing}
\vspace{-2mm}
\resizebox{\textwidth}{!}{%
\setlength{\tabcolsep}{6pt}
\begin{tabular}{l l cccc ccccc}
\toprule
& & \multicolumn{4}{c}{\textbf{In-domain}} & \multicolumn{5}{c}{\textbf{Out-of-domain}} \\
\cmidrule(lr){3-6} \cmidrule(lr){7-11}
& \textbf{Training Topics} & \textbf{MMLU} & \textbf{HellaSwag} & \textbf{GSM8K}$^\ast$ & \textbf{Avg.} & \textbf{GPQA} & \textbf{ARC-C} & \textbf{MATH-500}$^\ast$ & \textbf{Winogrande} & \textbf{Avg.} \\
\midrule
\multicolumn{11}{l}{\textit{Gemma 4 E2B}} \\
& Base & 37.5 & 48.7 & 72.7 & 53.0 & 28.8 & 39.1 & 56.5 & 50.0 & 43.6 \\
\multirow{-2}{*}{\textit{Reference}} & + Single domain & \textit{47.7}\subpos{10.2} & \textit{56.2}\subpos{7.5} & \textit{80.9}\subpos{8.2} & \textit{61.6}\subpos{8.6} &  &  &  &  &  \\
\cmidrule(l){2-11}
& + MMLU + HellaSwag & 46.9\subpos{9.4} & \textbf{55.6}\subpos{6.9} & 74.3\subpos{1.6} & 58.9\subpos{5.9} & \underline{30.3}\subpos{1.5} & \underline{44.9}\subpos{5.8} & 48.0\subneg{-8.5} & \underline{51.8}\subpos{1.8} & 43.8\subpos{0.2} \\
& + HellaSwag + GSM8K & 42.6\subpos{5.1} & 54.6\subpos{5.9} & 77.0\subpos{4.3} & 58.1\subpos{5.1} & \textbf{31.3}\subpos{2.5} & 43.0\subpos{3.9} & \underline{55.8}\subneg{-0.7} & \underline{51.8}\subpos{1.8} & \underline{45.5}\subpos{1.9} \\
& + MMLU + GSM8K & \textbf{48.8}\subpos{11.3} & 50.8\subpos{2.1} & \textbf{78.5}\subpos{5.8} & \underline{59.4}\subpos{6.4} & 28.8\subneu & 44.5\subpos{5.4} & \textbf{64.1}\subpos{7.6} & \textbf{52.7}\subpos{2.7} & \textbf{47.5}\subpos{3.9} \\
\rowcolor{gray!12}
\multirow{-4}{*}{\textit{Mixtures}} & \textbf{+ All three} & \underline{48.4}\subpos{10.9} & \underline{55.4}\subpos{6.7} & \underline{77.9}\subpos{5.2} & \textbf{60.6}\subpos{7.6} & 29.3\subpos{0.5} & \textbf{45.3}\subpos{6.2} & 51.6\subneg{-4.9} & 51.2\subpos{1.2} & 44.4\subpos{0.8} \\
\bottomrule
\multicolumn{11}{l}{\footnotesize $^\ast$Evaluated with thinking. All other benchmarks are evaluated without thinking.} \\
\end{tabular}%
}
\end{table*}

\textbf{Multi-Domain Curriculum \& Domain Generalization.}
Table~\ref{tab:domain_mixing} evaluates whether OPST can self-train on several domains at once by combining the topic descriptors of two or all three in-domain benchmarks. Training on all three domains improves the in-domain average by $7.6$ points, the largest gain among the mixtures, and comes within $1.0$ point of single-domain training, indicating little interference between domains. Out-of-domain retention is mixed: the average improves by $0.8$ points, but MATH-500 regresses by $4.9$ points. Combining MMLU with GSM8K topics improves MMLU by $11.3$ points, more than MMLU topics alone ($+10.2$), and yields the best out-of-domain average ($+3.9$), suggesting positive transfer from mathematical reasoning. Even without any MMLU topics, MMLU improves by $5.1$ points, so part of the gain is not specific to the seeded domain.

\subsection{Multimodal Self-Training: Visual OPST}
\label{sec:exp_visual}

To evaluate whether our self-training framework extends beyond text-only reasoning, we apply Visual OPST to Gemma 4 E2B and E4B vision-language models~\citep{team2026gemma4} using unlabeled images sampled from OpenImages V4~\citep{kuznetsova2020openimagesv4}. Table~\ref{tab:visual_results} reports performance on visual question answering (VQAv2~\citep{goyal2017vqav2}) and object hallucination detection (POPE~\citep{li2023pope}) %and object counting (CountBench~\citep{paiss2023teaching}).

\begin{table*}[t]
\centering
\small
\caption{\textbf{Multimodal results.} Performance comparison on vision-language benchmarks for Gemma 4 models self-trained on unlabeled images only.}
\label{tab:visual_results}
\vspace{-2mm}
\resizebox{\ifdim\width>\textwidth\textwidth\else\width\fi}{!}{%
\setlength{\tabcolsep}{6pt}
\begin{tabular}{l l cc c ccc}
\toprule
& & \multicolumn{2}{c}{\textbf{Privileged Info}} & \textbf{VQAv2} & \multicolumn{3}{c}{\textbf{POPE}} \\
\cmidrule(lr){3-4} \cmidrule(lr){5-5} \cmidrule(lr){6-8}
& \textbf{Method} & \textbf{Caption} & \textbf{Spatial} & \textbf{Acc.\,($\uparrow$)} & \textbf{F1\,($\uparrow$)} & \textbf{Acc.\,($\uparrow$)} & \textbf{Rec.\,($\uparrow$)} \\
\midrule
\multicolumn{8}{l}{\textit{Gemma 4 E2B}} \\
& Base & & & 49.1 & 82.2 & 83.2 & 72.9 \\
\cmidrule(l){2-8}
& + Visual OPST & $\times$ & $\checkmark$ & 55.7\subpos{6.6} & \textbf{84.7}\subpos{2.5} & \textbf{85.1}\subpos{1.9} & \textbf{77.4}\subpos{4.5} \\
& + Visual OPST & $\checkmark$ & $\times$ & 54.0\subpos{4.9} & 83.0\subpos{0.8} & 83.9\subpos{0.7} & 74.0\subpos{1.1} \\
\rowcolor{gray!12}
\multirow{-3}{*}{\textit{Ours}} & \textbf{+ Visual OPST} & $\checkmark$ & $\checkmark$ & \textbf{56.0}\subpos{6.9} & 84.6\subpos{2.4} & 85.0\subpos{1.8} & \textbf{77.4}\subpos{4.5} \\
\midrule
\multicolumn{8}{l}{\textit{Gemma 4 E4B}} \\
& Base & & & 55.9 & 84.8 & 85.2 & 78.0 \\
\cmidrule(l){2-8}
& + Visual OPST & $\times$ & $\checkmark$ & \textbf{59.6}\subpos{3.7} & \textbf{87.9}\subpos{3.1} & \textbf{87.4}\subpos{2.2} & \textbf{86.4}\subpos{8.5} \\
& + Visual OPST & $\checkmark$ & $\times$ & 58.0\subpos{2.2} & 87.8\subpos{2.9} & 87.2\subpos{2.0} & \textbf{86.4}\subpos{8.5} \\
\rowcolor{gray!12}
\multirow{-3}{*}{\textit{Ours}} & \textbf{+ Visual OPST} & $\checkmark$ & $\checkmark$ & 59.2\subpos{3.3} & 87.0\subpos{2.1} & 86.6\subpos{1.4} & 84.2\subpos{6.2} \\
\bottomrule
\end{tabular}%
}
\end{table*}

As shown in Table~\ref{tab:visual_results}, Visual OPST transfers to vision-language models using unlabeled images alone. Both forms of privileged information improve VQAv2 accuracy at 2B, by $6.6$ (spatial QA) and $4.9$ (scene caption) points. At 2B, combining them yields the largest VQAv2 gain ($+6.9$), while its POPE F1 gain ($+2.4$) matches spatial QA alone ($+2.5$). At 4B, spatial QA alone performs best on VQAv2 and POPE, so the two forms of privileged information are not consistently complementary. On POPE, the improvement comes mainly from recall ($+4.5$ points at 2B and $+6.2$ at 4B).%, while precision decreases slightly (from $94.2$ to $93.3$ at 2B and from $92.9$ to $90.0$ at 4B). The base models are conservative and deny many objects that are present in the image. After self-training, they miss fewer of these objects at a small cost in false positives.

\vspace{-2mm}
\section{Ablations}
\label{sec:ablations}

\begin{table*}[t]
\centering
\small
\caption{\textbf{Ablations on Gemma 4 E2B  on MMLU.} Default settings are highlighted in gray.} %In (a), filters are added cumulatively. In (b), \emph{Full} conditions the teacher on rationale and answer, and \emph{Placebo} on the privileged information of a different problem on the same topic. In (c), both filters are disabled and a fraction $p$ of candidate answers is replaced by a wrong answer; the base model scores 37.5.}
\label{tab:pipeline_ablations}
\vspace{-2mm}
\begin{minipage}[t]{0.34\textwidth}
\vspace{0pt}
\centering
\setlength{\tabcolsep}{6pt}
\begin{tabular}{lc}
\toprule
Sample selection & MMLU \\
\midrule
Base & 37.5 \\
+ OPST w/o verification & 42.7 \\
+ Verification & 45.4 \\
\rowcolor{gray!15}
\textbf{+ Consensus gate} & \textbf{47.7} \\
\bottomrule
\end{tabular}
\vspace{1.5mm}
\par\footnotesize \textbf{(a) Sample selection.}
\end{minipage}
\hfill
\begin{minipage}[t]{0.30\textwidth}
\vspace{0pt}
\centering
\setlength{\tabcolsep}{6pt}
\begin{tabular}{lc}
\toprule
Privileged info & MMLU \\
\midrule
None & 40.6 \\
Placebo & 40.2 \\
Answer only & 44.6 \\
Rationale only & 45.8 \\
\rowcolor{gray!15}
\textbf{Full} & 47.7 \\
\bottomrule
\end{tabular}
\vspace{1.5mm}
\par\footnotesize \textbf{(b) Privileged information.}
\end{minipage}
\hfill
\begin{minipage}[t]{0.28\textwidth}
\vspace{0pt}
\centering
\setlength{\tabcolsep}{6pt}
\begin{tabular}{ccc}
\toprule
$p$ & SFT & OPST \\
\midrule
0\% & 38.8 & 42.7 \\
25\% & 38.1 & 42.1 \\
50\% & 35.4 & 41.4 \\
75\% & 33.1 & 40.8 \\
\midrule
Drop & $-5.7$ & $-1.9$ \\
\bottomrule
\end{tabular}
\vspace{1.5mm}
\par\footnotesize \textbf{(c) Wrong candidate answers.}
\end{minipage}
\vspace{-5mm}
\end{table*}

\textbf{Quality Filtering.}
We ablate our quality filters by adding them incrementally, starting from OPST on all generated samples. Without filtering, OPST improves over the base model by $+5.2$ points (Table~\ref{tab:pipeline_ablations}a). Pseudo-verification adds $+2.7$ points and the consensus gate a further $+2.3$.

\textbf{Privileged Information.}
In Table~\ref{tab:pipeline_ablations}b, removing privileged information shrinks the gain over the base model from $10.2$ to $3.1$ points, and a placebo from a different problem on the same topic yields only $2.7$. The rationale alone recovers $8.3$ points and the answer alone $7.1$. The gain therefore stems from sample-specific privileged information, with the rationale carrying more signal than the answer; the residual gain without it likely reflects self-distillation towards the EMA teacher.

\textbf{Robustness to Wrong Answers.}
In Table~\ref{tab:pipeline_ablations}c, we disable both filters and replace a fraction $p$ of the candidate answers with wrong ones. At $p = 75\%$, OPST loses only $1.9$ points and remains $3.3$ points above the base model, whereas SFT loses $5.7$ points and falls $4.4$ points below it. Conditioning the teacher on the generated answer, rather than imitating it, therefore limits the influence of incorrect generations. Further ablations are provided in Appendix~\ref{subsec:more-ablations}.

\section{Conclusion}
\label{sec:conclusion}

We introduced On-Policy Self-Training (OPST), which post-trains a model on self-generated problems and privileged information, seeded only by coarse topic descriptors. We showed that conditioning the teacher on the generated answer, rather than training on it, lets the teacher fill in omitted reasoning steps and limits the influence of potentially noisy data. We found that OPST outperforms SFT and GRPO on the same generated data and methods that rely on curated questions, and extends to multiple domains and vision-language models, demonstrating that models can self-improve from self-generated data alone.

\paragraph{Limitations.}
OPST can only amplify existing knowledge, and since pseudo-verification measures self-consistency rather than correctness, consistent errors still enter training. Our topic descriptors are hand-chosen, and our experiments are limited to small models. Promising directions include label-free stopping criteria and tool- or retrieval-grounded verification.

\subsection*{AI use statement}

In this work, we used generative AI tools for implementing training infrastructure and experiment scripts and for feedback on the experimental design. We have not used generative AI tools for generating core research ideas, formulating hypotheses, or modifying evaluation benchmark data, and tasks involving human-subject data are not applicable to this work. Additionally, we used generative AI tools for drafting and editing parts of the manuscript, improving readability, checking and formatting tables, plotting utilities, and debugging LaTeX compilation. Open-weight generative models (Gemma~4) serve as the primary object of study and generate the synthetic training data within our proposed self-training pipeline. We have reviewed all AI-assisted work: all generated code was manually audited and tested for correctness by the authors, all mathematical formulations and experimental numbers were verified against primary training logs, all AI-suggested interpretations were checked against the results, and all references were checked against original publications. We take responsibility for the final content of this work, including text, claims, or artifacts produced with the aid of generative AI.

\bibliography{iclr2027_conference}
\bibliographystyle{iclr2027_conference}

\newpage

\appendix
\section{Appendix}
\subsection{Self-Distillation Objectives on Generated Data.}
Unlike a ground-truth reference solution, generated privileged information $z = (r, a)$ can be ambiguous: the rationale $r$ may miss important aspects, and the candidate answer $a$ may be partially or fully incorrect. Therefore, training a model on $z$ directly inherits these defects. SFT on the generated pairs maximizes
\begin{equation}
     \log \pi_\theta(z \mid x) = \sum_{j=1}^{|z|} \log \pi_\theta(z_j \mid x, z_{<j}),
\end{equation}
which supervises the model only on the tokens of $z$. Critically, it  treats every token as correct, including any potential omissions and errors. OPSD instead uses $z$ as context: at every state $y_{<t}$ the student visits, the teacher supplies a full next-token distribution, so supervision covers the entire student rollout rather than the $|z|$ tokens of the generation. Crucially, the teacher produces this rollout from its own knowledge under the exposure to the generated privileged information. This leads to a more nuanced, higher-quality supervision, since the teacher reweights the model's own next-token distribution by how consistent each continuation is with the privileged information. This has three consequences for generated data. First, $z$ need not spell out a step for the teacher to supervise it: tokens that move the rollout toward $a$ are upweighted, so the teacher completes reasoning that $r$ omits. Second, the target is expressed in the model's own distribution rather than in the surface form of $z$, so the student does not imitate the phrasing or formatting of the generation. Third, the prior limits the influence of a faulty $z$: an error shifts the target only to the extent that the model finds it plausible, and where the prior favors a different step, the teacher retains that step instead of copying the error. The student therefore learns from what the model can infer given $z$ as a hint, not from $z$ itself. Consequently, conditioning the teacher on $z$ yields dense, prior-grounded token-level supervision that filters generation noise, making on-policy self-distillation well suited for learning from purely synthetic data.

\subsection{Further Ablations}
\label{subsec:more-ablations}

To isolate the contribution of each design choice in OPST, we conduct systematic component ablations on Gemma~4 E2B across four core axes: the EMA update rate $\beta$, the synthetic dataset size generated per round, the number of outer self-training iterations $N$, and the evaluation target ($\pi_\phi$ vs.\ $\pi_\theta$). 

% ==================== ROW 1: (a), (b), (c) ====================
\begin{table*}[h]
\centering
\small
\caption{\textbf{OPST component ablations on Gemma~4 E2B evaluated on MMLU.} Default settings are highlighted in gray.}
\label{tab:component_ablations}
\vspace{2mm}

% ==================== ROW 1: (a), (b) ====================
\begin{minipage}[t]{0.46\textwidth}
\vspace{0pt}
\centering
\setlength{\tabcolsep}{10pt}
\begin{tabular}{lc}
\toprule
rate $\beta$ & MMLU \\
\midrule
1.00 (no EMA) & 38.1 \\
0.20 & 44.2 \\
\rowcolor{gray!15}
\textbf{0.05} & \textbf{47.7} \\
0.01 & 45.2 \\
\bottomrule
\end{tabular}
\vspace{1.5mm}
\par\footnotesize \textbf{(a) EMA update rate.}
\end{minipage}
\hfill
\begin{minipage}[t]{0.46\textwidth}
\vspace{0pt}
\centering
\setlength{\tabcolsep}{10pt}
\begin{tabular}{lc}
\toprule
samples / round & MMLU \\
\midrule
57 & 43.8 \\
171 & 47.2 \\
\rowcolor{gray!15}
\textbf{285} & \textbf{47.7} \\
570 & 46.9 \\
\bottomrule
\end{tabular}
\vspace{1.5mm}
\par\footnotesize \textbf{(b) Dataset size per round.}
\end{minipage}

\vspace{4mm}

% ==================== ROW 2: (c), (d) ====================
\begin{minipage}[t]{0.46\textwidth}
\vspace{0pt}
\centering
\setlength{\tabcolsep}{10pt}
\begin{tabular}{lc}
\toprule
rounds $N$ & MMLU \\
\midrule
0 (base IT) & 37.5 \\
3 & 43.1 \\
\rowcolor{gray!15}
\textbf{7} & \textbf{47.7} \\
12 & 46.2 \\
20 & 46.2 \\
\bottomrule
\end{tabular}
\vspace{1.5mm}
\par\footnotesize \textbf{(c) Self-training rounds.}
\end{minipage}
\hfill
\begin{minipage}[t]{0.46\textwidth}
\vspace{0pt}
\centering
\setlength{\tabcolsep}{10pt}
\begin{tabular}{lc}
\toprule
Eval checkpoint & MMLU \\
\midrule
Student ($\pi_\theta$) & 45.9 \\
\rowcolor{gray!15}
\textbf{EMA teacher ($\pi_\phi$)} & \textbf{47.7} \\
\bottomrule
\end{tabular}
\vspace{1.5mm}
\par\footnotesize \textbf{(d) Evaluation target.}
\end{minipage}
\end{table*}

\paragraph{EMA Update.}
The EMA teacher is meant to provide stable targets despite noisy updates on generated data (Section~\ref{sec:training}). We vary its rate $\beta$ (Table~\ref{tab:component_ablations}a). Without an EMA ($\beta = 1$), the teacher follows the student's current weights, and OPST barely improves over the base model ($38.1$ vs.\ $37.5$). Too much smoothing also slows progress: with $\beta = 0.01$, the teacher incorporates the student's improvements too slowly and reaches $45.2$, whereas $\beta = 0.05$ performs best ($47.7$).

\paragraph{Dataset Size per Round.}
The number of samples per round trades off the amount of data against how often the improving teacher regenerates it. We vary it from $57$ to $570$ samples (Table~\ref{tab:component_ablations}b). Accuracy increases up to $285$ samples and then saturates, decreasing slightly to $46.9$ at $570$ samples. Larger rounds therefore add generation cost without improving results.

\paragraph{Number of Rounds.}
Iterating lets student and teacher improve each other, but errors in the generated data can also accumulate over rounds (Section~\ref{sec:self_training_loop}). We evaluate the model after different numbers of rounds (Table~\ref{tab:component_ablations}c). Significant improvements emerge after the first three rounds ($+5.6$), and accuracy is highest after seven rounds ($47.7$). Further rounds bring no additional improvement, but performance remains stable at around $46$, well above the base model.

\paragraph{Evaluation Target.}
Because the EMA teacher averages over the student's recent updates, it may also be the better model to deploy. We compare the student and the EMA teacher at the end of training (Table~\ref{tab:component_ablations}d). The EMA teacher is $1.8$ points more accurate, consistent with the student's weights being affected by noise from recent batches that averaging removes.

\subsection{Variance Across Seeds}
\label{app:seeds}
To assess the stability of OPST, we repeat self-training on Gemma~4 E2B with three random seeds and report the mean and standard deviation in Table~\ref{tab:seed_variance}. The means correspond to the values reported in Table~\ref{tab:main_results}. Across both regimes, the standard deviation stays at or below 1.2 points and is therefore well below the gains of OPST over the base model. Without thinking, variance is low on all benchmarks. With thinking, it is highest on GPQA and GSM8K. On GPQA Diamond, which contains only 198 questions, a single question already corresponds to about 0.5 points. MMLU with thinking is the most stable benchmark (0.1), consistent with its large test set.

\begin{table}[h]
\centering
\small
\caption{\textbf{Variance across seeds.} Mean and standard deviation of OPST on Gemma~4 E2B over three seeds. Means correspond to Table~\ref{tab:main_results}.}
\label{tab:seed_variance}
\vspace{-2mm}
\setlength{\tabcolsep}{6pt}
\begin{tabular}{l ccccc}
\toprule
\textbf{Setting} & \textbf{MMLU} & \textbf{GPQA} & \textbf{GSM8K} & \textbf{HellaSwag} & \textbf{Average} \\
\midrule
w/o Think. & 47.7\std{0.8} & 32.8\std{1.0} & -- & 56.2\std{0.8} & 45.6\std{0.3} \\
w/ Think.  & 71.1\std{0.1} & 34.3\std{1.0} & 80.9\std{1.2} & 65.6\std{0.5} & 63.0\std{0.4} \\ % TODO: avg std approximated under independence; recompute from per-seed averages
\bottomrule
\end{tabular}
\end{table}

\subsection{Generalization to Qwen}
\label{app:qwen}
To test whether OPST is specific to the Gemma~4 model family, we additionally apply it to Qwen3-4B-Base \citep{yang2025qwen3}. Unlike the instruction-tuned Gemma~4 models used in the main experiments, Qwen3-4B-Base is a pretrained model without instruction tuning. We use the same topic descriptors, prompt templates, and hyperparameters as for Gemma~4 E2B % TODO: verify
and evaluate with the protocol described in Appendix~\ref{app:eval}. Table~\ref{tab:qwen_results} reports the results. OPST improves all four benchmarks and raises the average accuracy by $6.0$ points ($66.7 \to 72.7$). The largest gain occurs on GSM8K ($+13.3$), followed by GPQA ($+5.1$), MMLU ($+3.1$), and HellaSwag ($+2.5$). Since the multiple-choice benchmarks are scored by log-likelihood over the answer options, the gains on MMLU, GPQA, and HellaSwag do not stem from improved answer formatting. OPST therefore transfers both to a different model family and to a model that has not undergone instruction tuning.

\begin{table}[t]
\centering
\small
\caption{\textbf{Results on Qwen3-4B-Base.} Performance of OPST applied to a pretrained Qwen3 base model, evaluated with the protocol described in Appendix~\ref{app:eval}.}
\label{tab:qwen_results}
\vspace{-2mm}
\setlength{\tabcolsep}{6pt}
\begin{tabular}{l ccccc}
\toprule
\textbf{Method} & \textbf{MMLU} & \textbf{GPQA} & \textbf{GSM8K} & \textbf{HellaSwag} & \textbf{Average} \\
\midrule
\multicolumn{6}{l}{\textit{Qwen3-4B-Base}} \\
Base & 72.0 & 32.3 & 79.7 & 82.8 & 66.7 \\
\rowcolor{gray!12}
\textbf{+ OPST (Ours)} & \textbf{75.1}\subpos{3.1} & \textbf{37.4}\subpos{5.1} & \textbf{93.0}\subpos{13.3} & \textbf{85.4}\subpos{2.5} & \textbf{72.7}\subpos{6.0} \\
\bottomrule
\end{tabular}
\end{table}

\subsection{OPST Algorithm}
\label{app:algorithm}

Algorithm~\ref{alg:opst} summarizes the complete OPST training procedure. Both the student $\pi_\theta$ and the privileged teacher $\pi_\phi$ are initialized from the same weights $\theta_0$, so training requires no external teacher model. Each of the $N$ rounds consists of two stages: (1)~a fresh curriculum of problems $x$ with privileged information $z$ is synthesized from the topic set $\mathcal{T}$ and filtered via pseudo-verification and the consensus gate (Sections~\ref{sec:generation} and~\ref{sec:verification}), and (2)~the student is trained on its own on-policy rollouts $y \sim \pi_\theta(\cdot \mid x)$ by minimizing $\mathcal{L}_{\text{OPST}}$ (Eq.~\ref{eq:opst_loss}) against the teacher conditioned on $(x, z)$. After every update, the teacher tracks the student through an exponential moving average with rate $\beta$. The teacher therefore improves alongside the student while remaining a stable distillation target. After the final round, the EMA teacher $\pi_\phi$ is returned as the final model.

\begin{algorithm}[h!]
\caption{On-Policy Self-Training (OPST)}
\label{alg:opst}
\begin{algorithmic}[1]
\Require topics $\mathcal{T}$, initial weights $\theta_0$, rounds $N$, EMA rate $\beta$
\State $\theta \gets \theta_0$, \ $\phi \gets \theta_0$
\For{$n = 1, \dots, N$}
    \State Generate $(x_i, z_i) \sim \pi_{\text{gen}}(\cdot \mid \tau)$ for $\tau \in \mathcal{T}$ \Comment{\S\ref{sec:generation}}
    \State Pseudo-verification and consensus gate; discard noisy, trivial, and duplicate samples \Comment{Section~\ref{sec:verification}}
    \For{each mini-batch $\mathcal{B}$}
        \State Sample rollouts $y_i \sim \pi_\theta(\cdot \mid x_i)$ for all samples in $\mathcal{B}$
        \State Update $\theta$ with $\nabla_\theta \mathcal{L}_{\text{OPST}}(\theta; \mathcal{B})$ \Comment{Eq.~\ref{eq:opst_loss}}
        \State $\phi \gets (1-\beta)\,\phi + \beta\,\theta$ 
    \EndFor
\EndFor
\State \Return $\pi_\phi$
\end{algorithmic}
\end{algorithm}

\subsection{Evaluation Protocol} \label{app:eval}
We follow the task formulations of the EleutherAI \texttt{lm-evaluation-harness}~\citep{eval-harness} and adapt them to the Gemma~4 chat template. Without thinking, we score all multiple-choice benchmarks, i.e., MMLU, GPQA Diamond, HellaSwag, ARC-Challenge, and Winogrande, by log-likelihood over the closed set of answer options. No answer is therefore parsed from generated text. We use the harness defaults, i.e., 5-shot prompting for MMLU, zero-shot prompting for the remaining multiple-choice benchmarks, and length-normalized accuracy where the harness specifies it. GSM8K is evaluated with 8-shot chain-of-thought prompting \citep{wei2022chain}.

With thinking, multiple-choice answers are parsed from the final \texttt{Answer: X} line. GSM8K is graded by exact match of the final numerical answer, and MATH-500 by symbolic equivalence of the final \texttt{\textbackslash boxed\{\}} expression using SymPy. For all models and benchmarks, both before and after self-training, the rate of parsing failures and invalid answers is 0\%. Performance differences in the thinking regime therefore do not stem from changes in answer formatting. All remaining settings follow the harness defaults.

\subsection{OPST Prompt Templates}
\label{app:prompts}

We list the prompts used for problem synthesis, student rollouts, and teacher conditioning. Synthesis prompts are shown in gray, student prompts in blue, and teacher prompts in orange, with placeholders filled at runtime shown in \textcolor{teal!70!black}{\texttt{\textbf{\{color\}}}}. With thinking, the system turn \lstinline[style=prompt,basicstyle=\ttfamily]!<|turn>system\n<|think|><turn|>\n! is prepended to both student and teacher prompts.

\paragraph{Problem Synthesis.}
The generator receives a topic descriptor, e.g., \texttt{"college\_mathematics"} for MMLU or \texttt{"rate, time, and distance word problems"} for GSM8K. It returns a question $x$ together with a rationale $r$ and a candidate answer $a$. We use a multiple-choice template for MMLU, GPQA, and HellaSwag, and an open-ended template for GSM8K.

\begin{promptbox}{Synthesis Prompt (Multiple-Choice: MMLU, GPQA, HellaSwag)}
You are an expert exam writer creating challenging multiple-choice questions on
the topic of "{topic}".
 
Write ONE novel, self-contained, graduate-level multiple-choice question testing
deep conceptual understanding or multi-step reasoning in "{topic}". Provide
exactly four plausible answer choices labeled (A), (B), (C), and (D), followed
by a rigorous step-by-step solution and the final correct option letter.
 
Use the following exact format:
Question: <question text>
(A) <choice A>
(B) <choice B>
(C) <choice C>
(D) <choice D>
Solution: <step-by-step derivation explaining why the correct choice holds>
Answer: <single letter A, B, C, or D>
\end{promptbox}

\begin{promptbox}{Synthesis Prompt (Open-Ended: GSM8K)}
You are an expert mathematics curriculum designer creating multi-step word
problems on the topic of "{topic}".
 
Write ONE novel, self-contained arithmetic or algebraic word problem about
"{topic}" that requires 3 to 5 reasoning steps to solve. Ensure all numerical
quantities are consistent and yield an exact numerical answer. After the
question, provide a complete step-by-step worked solution and the final number.
 
Use the following exact format:
Question: <word problem text>
Solution: <step-by-step mathematical derivation>
Answer: <final numerical value>
\end{promptbox}

\paragraph{Student Prompts.}
The student receives only the question $x$.

\begin{promptbox}[blue]{Student Prompt (Multiple-Choice)}
Solve the following multiple-choice question step by step. After your reasoning,
state your final choice on a new line in the exact format "Answer: X" where X is
A, B, C, or D.
 
Question: {question}
(A) {choice_A}
(B) {choice_B}
(C) {choice_C}
(D) {choice_D}
\end{promptbox}

\begin{promptbox}[blue]{Student Prompt (Open-Ended)}
Solve the following math word problem step by step. Show all intermediate
calculations and conclude with your final numerical result in the exact format
"Answer: <number>".
 
Question: {question}
\end{promptbox}

\paragraph{Teacher Prompts.}
The teacher receives the question together with the privileged information $z = (r, a)$ and is instructed to solve the problem without referring to it. If the generated rationale is incomplete, the teacher first completes it, conditioned on the question and the candidate answer.

\begin{promptbox}[orange]{Rationale Completion Prompt}
Question: {question}
{choices_block}
 
The potentially correct answer to this problem is ({gen_answer}).
Write a clear, rigorous step-by-step derivation reasoning about ({gen_answer}).
Write the explanation as a natural, self-contained solution from first
principles. Do NOT mention that the correct answer was given to you in advance.
\end{promptbox}

\begin{promptbox}[orange]{Teacher Prompt}
You are an expert tutor solving the problem below. You are provided with a
potential reference solution and final answer as privileged background context.
Use this privileged information to guide a clear, rigorous, step-by-step
derivation of the problem from first principles. Do NOT mention the reference
solution or state that the answer was provided to you; present the reasoning as
a direct solution to the question.
 
### Privileged Reference Context:
Reference Rationale: {privileged_rationale}
Potential Answer: {gen_answer}
 
### Problem to Solve:
Question: {question}
{choices_block}
 
Provide your step-by-step reasoning and conclude with "Answer: ".
\end{promptbox}

\subsection{Visual OPST Prompt Templates}
\label{app:visual_prompts}

We list the prompts used in Visual OPST for visual question synthesis, privileged information extraction, visual verification, and student--teacher distillation. For all multimodal turns, the image token \lstinline[style=prompt,basicstyle=\ttfamily]!<|image|>\n! is prepended to the user turn \lstinline[style=prompt,basicstyle=\ttfamily]!<|turn>user\n...\n<turn|>\n<|turn>model\n!.

\paragraph{Visual Problem Synthesis.}
Given an unlabeled image $I \in \mathcal{I}$ and a visual topic descriptor $\tau \in \mathcal{T}_{\text{vis}}$, the generator synthesizes \texttt{\{n\}} question--answer pairs. We use a dedicated prompt for object counting and a shared template for the remaining visual topics.

\begin{promptbox}{Visual Synthesis Prompt (Object Counting)}
Look at this image carefully. Generate {n} counting questions about objects in
the image. Include some objects that appear multiple times and some that do not
appear at all (count 0).
Format EXACTLY as:
Q: <question>
A: <number>

Generate {n} different questions.
\end{promptbox}

\begin{promptbox}{Visual Synthesis Prompt (Attributes and Spatial Relations)}
Look at this image carefully. Generate {n} questions about {topic_description}.
Format EXACTLY as:
Q: <question>
A: <short answer>

Generate {n} different questions.
\end{promptbox}

\noindent where \texttt{\{topic\_description\}} is instantiated per topic $\tau \in \mathcal{T}_{\text{vis}}$ as:
\begin{itemize}
    \item \texttt{color\_attributes}: \emph{``colors, sizes, shapes, or visual attributes of objects''}
    \item \texttt{spatial\_relations}: \emph{``spatial relationships between objects (left/right, above/below, in front/behind, near/far)''}
\end{itemize}

\paragraph{Privileged Information.}
To construct the privileged information $z$, the generator either writes a detailed scene caption for $I$ or detects the objects in the scene with bounding boxes. From pairs of detected boxes, we derive grounded spatial relations between objects from their box centers (\texttt{"left of"}, \texttt{"right of"}, \texttt{"above"}, \texttt{"below"}).

\begin{promptbox}{Privileged Scene Caption Prompt}
Describe this image in detail, including the main objects, their attributes, and
their spatial arrangement.
\end{promptbox}

\begin{promptbox}{Object Detection Prompt}
Detect all objects in this image. For each object, provide the name and bounding
box coordinates.
Format each detection as: name: [x1, y1, x2, y2]
Coordinates are normalized 0-1000.
List up to 10 objects.
\end{promptbox}

\paragraph{Visual Verification.}
To compute $v_{\text{aug}}(I, x, z)$, the generator answers $x$ on each of the $K$ augmented views $\mathcal{A}_k(I)$ using a concise VQA prompt.

\begin{promptbox}{Visual Verification Prompt}
{question}
Answer concisely in 1-5 words.
\end{promptbox}

\paragraph{Student and Teacher Prompts.}
During distillation, the student receives the image $I$ and the question $x$, while the teacher additionally receives the privileged information $z$ prepended to $x$.

\begin{promptbox}[blue]{Visual Student Prompt}
<|image|>
{question}
\end{promptbox}

\begin{promptbox}[orange]{Visual Teacher Prompt}
<|image|>
Image description: {privileged_caption}  Objects present: {detected_objects}
{question}
\end{promptbox}

\newpage

\end{document}